\documentclass[a4paper,twoside]{article}

\usepackage{epsfig}
\usepackage{subfigure}
\usepackage{calc}
\usepackage{amssymb}
\usepackage{amstext}
\usepackage{amsmath}
\usepackage{amsthm}
\usepackage{multicol}
\usepackage{pslatex}
\usepackage{graphicx}
\usepackage{booktabs}
\usepackage[table]{xcolor}
\usepackage{tikz}
\usetikzlibrary{calc}
\usepackage{natbib}
\usepackage{balance}

\newcommand{\tikzmark}[1]{\tikz[overlay,remember picture] \node (#1) {};}

\newcommand*{\DrawArrow}[3][]{%
    \begin{tikzpicture}[overlay,remember picture]
        \draw [very thick, -stealth, #1] ($(#2)+(0.25em,-0.3ex)$) to ($(#3)+(0.25em,4ex)$);
    \end{tikzpicture}%
}%

\usepackage{SCITEPRESS}

\begin{document}

\title{Combining Bayesian Approaches and Evolutionary Techniques for the Inference of Breast Cancer Networks}

\author{\authorname{Stefano Beretta\sup{1}, Mauro Castelli\sup{2}, Ivo Gon\c{c}alves\sup{2}, Ivan Merelli\sup{3} and Daniele Ramazzotti\sup{4}}
	\affiliation{\sup{1}DISCo, Universit\'a degli Studi di Milano Bicocca. 20126 Milano, Italy}
	\affiliation{\sup{2}NOVA IMS, Universidade Nova de Lisboa. 1070-312 Lisboa, Portugal}
	\affiliation{\sup{3}Ist. di Tecnologie Biomediche, Consiglio Nazionale delle Ricerche, Segrate, Italy}
	\affiliation{\sup{4}Department of Pathology, Stanford University}
	\email{stefano.beretta@disco.unimib.it, \{mcastelli,igoncalves\}@novaims.unl.pt, ivan.merelli@itb.cnr.it, daniele.ramazzotti@stanford.edu}
}


\abstract{Gene and protein networks are very important to model complex
large-scale systems in molecular biology.
Inferring or reverse‐engineering such networks can be defined as the
process of identifying gene/protein interactions from experimental
data through computational analysis. However, this task is typically
complicated by the enormously large scale of the unknowns in a rather
small sample size. Furthermore, when the goal is to study causal relationships 
within the network, tools capable of
overcoming the limitations of correlation networks are required. 
In this work, we make use of Bayesian Graphical Models to attach this problem 
and, specifically, we perform a comparative study of different state-of-the-art 
heuristics, analyzing their performance in inferring the structure of the Bayesian Network
from breast cancer data.
}

\onecolumn \maketitle \normalsize \vfill

\section{\uppercase{Introduction}}
\label{sec:intro}
\noindent
Molecular networks are essential for every biological process, since genes and proteins are able to carry out their function only in precisely regulated pathways. For this reason, data-driven learning of regulatory connections in molecular networks has long been a key topic in computational biology~\citep{1}. The general problem is to infer, or reverse-engineer, from gene or protein expression data, the regulatory interactions among these biological entities using computational algorithms.

In this context, despite correlation networks are widely used for gene expression and proteomic data analysis, it is known that correlations not only confound direct and indirect associations, but also provide no means to distinguish between cause and effect. For causal analysis the inference of a directed graphical model is typically  required. However, this task is rather difficult due to multiple theoretical and practical reasons, among which, but not limited to, the course of dimensionality~\citep{3}.

Therefore, causal analysis requires tools capable of overcoming the limitations of correlation networks: much of the work in this area has focused on Bayesian Networks~\citep{3} or related regression models, such as systems of recursive equations or influence diagrams. All these models describe causal relations by an underlying directed acyclic graph (DAG). Nevertheless, it remains unclear whether causal, rather than merely correlational, relationships in molecular networks can be inferred in complex biological settings.

Moreover, the problem is typically complicated by the enormously large scale of the unknowns in a rather small sample size. Furthermore, data is prone to experimental defects and noisy readings, while many other biases can compromise the quality of the results. These complexities call for a heavy involvement of powerful mathematical models which play an increasingly important role in this research area~\citep{4}.
In order to assess the ability of different tools to learn causal networks, the Dialogue for Reverse Engineering Assessment and Methods (DREAM) project has run several challenges focused on network inferences~\citep{2}. In particular, we focused on (sub)-challenge 8.1 concerning Human Protein Networks (HPN) in cancer cell lines, which is about the inference of causal signalling pathways using time-course data with perturbations on network nodes. This sub-challenge was split into two independent parts, concerning Breast Cancer proteomic data and in silico data.

Different types of models, such as directed graphs, Boolean networks~\citep{6}, Bayesian Graphical Models~\citep{7}, and various differential models have been used to describe gene regulations at various levels of detail and complexity. The choice of the model is often determined by how much information it tries to capture, taking into account that the more information a model attempts to infer, the more parameters are needed to learn it, and the more complex the overall approach becomes. 
Specifically, researchers have paid great attention to Bayesian Networks, which can compactly model dependency relationships between variables relying on probabilistic measures. Since gene expression experiments are subject to many measurement errors, the use of statistical methods is expected to be effective for extracting useful information from such noisy data. Friedman et al.~\citep{9} proposed both discrete and continuous Bayesian network models relying on linear regression for inferring gene networks. Imoto et al.~\citep{10} succeeded in employing non-parametric regressions for capturing even non-linear relationships between genes.

In this work, we perform a comparative study of different heuristics at the state-of-the-art to perform the task of inferring the structure of a Bayesian network from breast cancer data. The paper is structured as follows: Section~\ref{sec:back} provides a background of the biological problem under exam; Section~\ref{sec:meth} gives a formal definition of the problem addressed in this study, along with a description of the different computational and statistical machineries that we are adopting, and of the input data. Afterwards, the results of the described methods on real and simulated data are presented and discussed in Section~\ref{sec:result}. Section~\ref{sec:concl} concludes the paper and suggests avenues for future research.

\section{\uppercase{Biological Background}}
\label{sec:back}

\noindent
Many biological processes are carried out by interactions between proteins, RNA, and DNA. Cells respond to their environment by activating signalling networks that trigger processes such as growth, survival, apoptosis (programmed cell death), and migration. Post-translational modifications, notably phosphorylation, play a key role in these signalling events. 
In cancer cells, signalling networks frequently become compromised, leading to abnormal behaviours and responses to external stimuli. Endogenous signal transduction in cancer cells is systematically disturbed to redirect the cellular decisions from differentiation and apoptosis to proliferation and, later, invasion. Cancer cells acquire their malignancy through accumulation of advantageous gene mutations by which the necessary steps to malignancy are obtained. These selfish adaptations to independence can be described as a result from an evolutionary process of diversity and selection~\citep{11}.

Many current and emerging cancer treatments are designed to block nodes in signalling networks, thereby altering signalling cascades. Although there is a wealth of literature describing canonical cell signalling networks, little is known about exactly how these networks operate in different cancer cells. Advancing our understanding of how these networks are deregulated across cancer cells will ultimately lead to more effective treatment strategies for patients.

Recently, high-throughput analysis enabled the possibility to obtain genome-wide information, such as mRNA expressions, protein-protein interactions, protein localizations and so on. A lot of attention has been dedicated on developing computational methods for extracting valuable information of molecular networks from such various types of genomic data. 

Currently, statistical models for estimating gene regulatory networks from genomic data are mainly based on expression data from DNA microarrays or RNA-seq experiments. However, since information from these approaches is limited by their quality, noise and experimental errors, sophisticated mathematical approaches are necessary for estimating gene regulatory networks accurately.

On the other hand, protein-protein interaction networks are mainly constructed relying on observed protein-protein interaction data, using approaches such the two hybrid assays, tandem affinity purification experiments and, more recently, protein arrays. However, protein-protein interaction data often contains some errors, making even more difficult to construct comprehensive protein-protein interaction networks from these interaction data alone.

\section{\uppercase{Methods}}
\label{sec:meth}
\noindent
A Bayesian Network (BN) is a statistical graphical model that represents a joint distribution over $n$ random variables and encodes it by means of a direct acyclic graph (DAG) depicting the $n$ nodes referring to the variables. More formally, we define a BN as a direct acyclic graph $G=(V,E)$, where $V$ is the set containing the $n$ random variables and $E$ is the set of the directed arcs over them, representing any conditional dependence among the variables \citep{koller_book}. 

In this work, we make use of such graphical tool to model a protein network $G_p$ (being a direct acyclic graph), whose structure (i.e., the nodes and arcs in the model) maximizes the likelihood, given the observed data on which we make the inference. Moreover, we define this task as an optimization problem where, for a set of observations $D$, we aim at maximizing the likelihood of observing the data given a specific model $G_p$, which we define as
\[
\mathcal{LL}(G_p,D) = \prod_{d \in D} P(d|G_p)\, ,
\]
that is the product of the conditional probabilities given each observation $d \in D$. 

Practically, however, there is a well-known issue when learning the network structure by maximizing the likelihood function. In fact, for any arbitrary set of data, the most likely graph is usually very connected, since adding an edge typically can only increase the likelihood of the data, hence leading to overfitting. To try to reduce this problem, the likelihood is almost always adjusted by means of a regularization term that penalizes the complexity of the model \citep{koller_book}. 

We also observe that, regardless of the adopted approach and likelihood score, the main issue to infer the structure of a BN is the huge search space of the valid solutions, which makes this a well known NP-hard problem and, therefore, one will need to make use of heuristics to perform such inference \citep{koller_book}. 

In this work, we compare different heuristics search algorithms along with various regularizations for the likelihood score. In Table~\ref{table:comb} we present a list of combinations of the adopted techniques, which are described in details in the subsequent sections. 

\begin{table}[ht]
\centering
\begin{tabular}{l | ccc}
\toprule
Heuristic Search Algorithm & \multicolumn{3}{l}{Regularizators} \\
\midrule
Hill Climbing (HC) & loglik & AIC & BIC \\
Tabu Search (TB) & loglik & AIC & BIC \\
Genetic Algoritms (GA) & loglik & AIC & BIC \\
\bottomrule
\end{tabular}
\caption {Combinations of the different heuristics and regularization
approaches used in this work.}
\label{table:comb}
\end{table}

Here we employ three different and well-known evolutionary methods to solve the previously mentioned optimization problem, that is to reconstruct the Bayesian network w.r.t.~to a specific regularization score.
In the rest of this section we briefly describe each method and also the considered regularizators.

\subsection{Hill climbing}
\label{sec:hill}
\noindent

Hill Climbing (HC) is one of the simplest iterative techniques that have been proposed for solving optimization problems. While HC consists of a simple and intuitive sequence of steps, it is a good search technique to be used as a baseline for comparing the performance of more advanced optimization techniques.
%

Hill climbing shares with other techniques (like simulated annealing~\citep{hwang1988simulated} and tabu search~\citep{glover1989tabu}) the concept of neighbourhood. Search methods based on this latter concept are iterative procedures in which a neighbourhood $N(i)$ is defined for each feasible solution $i$, and the next solution $j$ is searched among the solutions in $N(i)$. Hence, the neighbourhood is a function $N:S\rightarrow 2^S$ that assigns at each solution in the search space $S$ a (non-empty) subset of $S$.
In our case, every solution is modelled as an adjacency matrix, where an entry $[i,j]$ is $1$ if in the current solution an arc is present from node $i$ to node $j$, and is $0$ otherwise.

The sequence of steps of the hill climbing algorithm, for a minimization
problem w.r.t.~a given objective function $f$, are the following:
\begin{enumerate}
\item choose an initial solution $i$ in $S$;
\item find the best solution $j$ in $N(i)$ (i.e., the solution $j$ such that $f(j)\leq f(k)$ for every $k$ in $N(i)$;
\item if $f(j) > f(i)$, then stop; else set $i=j$ and go to Step 2.
\end{enumerate}

%
To counteract the main limitation of hill climbing (i.e., getting trapped in a local optimum), more advanced neighbourhood search methods have been defined. The following section presents the Tabu Search method, a popular and effective optimization technique that uses the concept of ``memory''.

\subsection{Tabu search}
\label{sec:tabu}

\noindent
As described in the original work of Glover~\citep{glover1989tabu}, Tabu Search (TS) is a meta-heuristic that guides a local heuristic search procedure to explore the solution space beyond local optimality. One of the main components of this method is the use of an adaptive memory, which creates a more flexible search behaviour. Memory-based strategies are therefore the main feature of TS approaches, founded on a quest for ``integrating principles'', by which alternative forms of memory are appropriately combined with effective strategies for exploiting them. 

\emph{Tabus} are one of the distinctive elements of TS when compared to hill climbing or other local search methods. The main idea in considering \emph{tabus} is to prevent cycling when moving away from local optima through non-improving moves. When this situation occurs, something needs to be done to prevent the search from tracing back its steps to where it came from. This is achieved by declaring \emph{tabu} (disallowing) moves that reverse the effect of recent moves. For instance, let us consider a problem where solutions are binary strings of a prefixed length and the neighbourhood of a solution $i$ consists of the solutions that can be obtained from $i$ by flipping only one of its bits. In this scenario, if a solution $j$ has been obtained from a solution $i$ by changing one bit $b$, it is possible to declare a \emph{tabu} to avoid to flip back the same bit $b$ of $j$ for some number of iterations (this number is called the tabu tenure of the move). \emph{Tabus} are also useful to help the search move away from previously visited portions of the search space and, thus, perform more extensive exploration.



The basic TS algorithm is reported, considering the minimization of the objective function $f$, as follows:
\begin{enumerate}
 \item randomly select an initial solution $i$ in the search space $S$, and set $i^* = i$ and $k=0$, where $i^*$ is the best solution so far and $k$ the iteration counter;
 \item set $k=k+1$ and generate the subset $V$ of the \emph{admissible} neighbourhood solutions of $i$ (i.e., non-\emph{tabu} or allowed by aspiration);
\item choose the best $j$ in $V$ and set $i = j$;
\item if $f(i) < f(i^*)$, then set $i^* = i$;
\item update \emph{tabu} and aspiration conditions;
\item if a stopping condition is met then stop; else go to Step 2.
\end{enumerate}

Commonly used conditions to end the algorithm are when the number of iterations ($K$) is larger than the maximum number of allowed iterations, or if no changes to the best solution have been performed in the last $N$ iterations (as in our tests).

\subsection{Genetic Algorithm}
\label{sec:GA}
\noindent

Genetic Algorithms (GAs) are a class of computational models that mimic the process of natural evolution~\citep{goldberg1988genetic}. GAs are often considered as function optimizers although the range of problems to which genetic algorithms have been applied is quite broad. Although different variants exist, most of the methods called ``GAs'' have at least the following elements in common: populations of chromosomes, selection according to a fitness function, crossover to produce new offspring, and random mutation of new offspring.

One of the most important issues when using the GAs to solve an optimization problem is the way to encode the candidate solutions, that is the individuals in the population, and also the genetic operators (crossover and mutation).
Since, this aspect strongly depends on the specific problem, here we describe how GAs have been used to build a Bayesian Network.
A candidate solution is represented as a string $s$ of length equal to $n^2$, being $n$ the number of nodes of the network. Each position $s[i]$ can be either $0$ or $1$, and the information represents the existence of a connection among node $i/n$ and node $i\%n$, where the $/$ operator denotes the integer division, while the $\%$ operators denotes the rest of the division between $i$ and $n$. As an example, $s[12]=1$ in a network with $10$ nodes means that there is a node between node 1 ($12/10$) and node 2 ($12\%10$). Nodes are numbered from $0$ to $n-1$.

To produce admissible solutions (i.e., in our domain a network without loops), it is fundamental to redefine the classical crossover and mutation operators.
More precisely, we developed a simple but efficient method that guarantees that crossover and mutation will produce Bayesian Networks without loops. To achieve this goal we associated to each solution two lists, called forward list and backward list. The two lists maintain, for each node $k$, the forward links (i.e., the set of nodes $\hat{k}$ for which a connection from $k$ to $\hat{k}$ exists) and the backward links (i.e., the set of nodes $\hat{k}$ for which a connection from $\hat{k}$ to $k$ exists).
By using these two linked lists it is simple to assess if a new connection between two nodes can be created. In detail, let us assume that the algorithm needs to evaluate whether it is possible to add a connection between nodes $k_1$ and $k_2$ (with $k_1$ being the origin and $k_2$ the destination node of the connection). In this scenario, it is necessary to iteratively scan all the elements in the backward list of $k_1$ and check if in their backward lists $k_2$ is present. In this case it would be impossible to create a connection between $k_1$ and $k_2$ without entering a loop in the structure of the network. In the same way, it is necessary to iteratively scan all the elements in the forward list of $k_2$ and check if in their forward lists $k_1$ is present. Also in this case, the creation of the connection from $k_1$ to $k_2$ will introduce a loop in the network.

Hence, the proposed crossover operator works as follows:
\begin{enumerate}
	\item choose two individuals $p_1$ and $p_2$ as parents, based on tournament selection;
	\item select a single crossover point for both the parents;
	\item for every locus $i$ before that point set $child_1[i]=p_1[i]$ and $child_2[i]=p_2[i]$;
	\item for every locus $i$ beyond that point for which $p_1[i]$ is equal to $p_2[i]$, set $child_1[i]=p_1[i]$ and $child_2[i]=p_2[i]$;
	\item for every locus $i$ beyond that point for which $p_1[i]$ is different from $p_2[i]$, do the following:
	\begin{itemize}
	\item if $p_2[i]=0$, then set $child_1[i]=0$ and set $child_2[i]=1$ if and only if it is possible to create a connection between node $i/n$ and node $i\%n$ (set $child_2[i]=0$ in the opposite case);
	\item if $p_1[i]=0$, then set $child_2[i]=0$ and set $child_1[i]=1$ if and only if it is possible to create a connection between node $i/n$ and node $i\%n$ (set $child_1[i]=0$ in the opposite case);
	\item  update the forward and the backward lists.
  \end{itemize}
\end{enumerate}

The mutation operator we proposed works as follows:
\begin{enumerate}
	\item for each locus $i$ of an individual $p$ generate a random number $r$ from a uniform distribution. If $r \leq p_m$ (where $p_m$ is the mutation probability) then select the locus $i$ for mutation;
	\item if $p[i]=1$, then set $p[i]=0$ and update the forward and backward lists;
	\item in the opposite case ($p[i]=0$), check if it is possible to create a connection between node $i/n$ and node $i\%n$. If the connection does not introduce a loop set $p[i]=1$ and update the data structures, else $p[i]$ will remain equal to $0$.
	\end{enumerate}

The genetic operators described above ensure that the constraint related to the absence of loops is always satisfied. Moreover, this allows the GA to avoid to reject a high number of individuals that do not respect the constraint. This will result in a beneficial effect on the execution time of the algorithm.


\subsection{Regularizators}
\label{sec:scores}

\noindent
As already mentioned, we make use of various likelihood scores as fitness functions for the inference of the network. Such scores, namely \emph{loglik}, \emph{AIC}, and \emph{BIC}, are implemented by using the \emph{bnlearn R package} \citep{bnlearn}. 

Specifically, we first considered the log-likelihood score (\emph{loglik}), that is the logarithm of the previously mentioned likelihood score. 
Then, as regularized log-likelihood scores, we used the Akaike Information Criterion (\emph{AIC}) \citep{aic_score} and the Bayesian Information Criterion (\emph{BIC}) \citep{bic_score}. 

To extend this scores in order to model continuous random variables, we adopt the multivariate Gaussian implementation of the log-likelihood score (see \citep{koller_book} for a formal definition of the scores and \citep{bnlearn} for the adopted implementation).

\section{\uppercase{Results}}
\label{sec:result}

\noindent
To assess the performance of the different approaches and regularizators, we have considered the HPN-DREAM breast cancer network inference challenge.
This challenge comprises three sub-challenges, and we focused on the first one (\emph{Sub-challenge 1}). This sub-challenge consists of two distinct parts: the first one (\emph{Sub-challenge 1A}) aims at inferring causal signalling networks using protein time-course data. The task spanned $32$ different contexts, each defined by a combination of $4$ cell lines and $8$ stimuli, which focus on networks with specific genetic and epigenetic background. Since for these datasets the real network is unknown, beside training data, further data (not used during the inference) are available to assess the causal validity of the inferred networks. The second part (\emph{Sub-challenge 1B}) comprises in silico data task and also focused on causal networks. Anyway, differently from the former one, the use of a-priori biology knowledge to design the network is not allowed. Since for this sub-challenge the protein network is known, the evaluation of the achieved results can be performed by directly comparing the computed network with the original one.


More in details, the datasets of Sub-challenge 1A (``real data'') were generated using Reverse Phase Protein Array (RPPA) quantitative proteomics technology. RPPA is a protein array designed as a micro- or nano-scaled dot-blot platform that allows the simultaneous measurement of protein expression levels in a large number of biological samples in a quantitative manner, when high-quality antibodies are available \citep{RPPA}.
This challenge focuses on about $45$ phosphoproteins (proteins phosphorylated at specific sites). Protein abundance may be influenced by multiple dynamical processes operating over multiple time-scales. This challenge does not focus on long-term changes over days (e.g.~rewiring of networks due to epigenetic changes brought about by perturbation), hence data comprises protein time-course data up to 4 hours after ligand stimulation.
Time-course data were acquired under $8$ ligand stimuli and inhibition of network nodes by one of $3$ inhibitors plus the vehicle control (cells were serum-starved and pre-treated with inhibitor prior to ligand stimulation).
The experiment was carried out on $4$ breast cancer cell lines (namely, BT20, BT549, MCF7, and UACC812), with abundance of the $\sim 45$ phosphoproteins measured at $7$ time points post-stimulus. Data are normalized protein abundance measurements on a linear scale.
Table~\ref{fig:tab-exp} shows the $32$ different processed datasets, obtained by each combination of cell/stimulus, and their compositions, which are the expression levels of the considered phosphoproteins with $4$ different inhibitors at $7$ consecutive time points.

\begin{table*}[ht]
\centering
\begin{tabular}{|c|c|c|c|c|c|c|c|c|}
\toprule
& Serum & PBS & NRG1 & Insulin & IGF1 & HGF & FGF1 & EGF \\
\midrule
BT20 &&&&&&&&\\
\midrule
BT549 &&&&&&&&\\
\midrule
MCF7 &&&&&&&&\\
\midrule
UACC812 &&&&\cellcolor{gray!50}\tikzmark{first}&&&&\\
\bottomrule
\end{tabular}

\vspace{8mm}

\begin{tabular}{|c|p{6mm}|p{6mm}|p{6mm}|p{6mm}|p{6mm}|p{6mm}|p{6mm}|}
\cmidrule{2-8}
\multicolumn{1}{c|}{}& \tikzmark{second}$0$m & $5$m & $15$m & $30$m & $1$h & $2$h & $4$h\\
\midrule
GSK690693 &&&&&&&\\
\midrule
GSK690693\_GSK1120212 &&&&&&&\\
\midrule
PD173074 &&&&&&&\\
\midrule
DMSO &&&&&&&\\
\midrule
\end{tabular}
\DrawArrow[black, out=-90, in=90]{first}{second}
\caption{The upper table highlights the $32$ combinations of cells/stimuli which constitute the processed ``real datasets''. The lower table represents the composition of a single dataset (UACC812/Insulin in the example), which contains the expression levels of the phosphoproteins with $4$ inhibitors at $7$ different time points.}
\label{fig:tab-exp}
\end{table*}


On the other hand, the in silico challenge aims to mimic the key aspects of the RPPA experimental set up and the characteristics of the proteomic data, but using a state-of-the-art dynamical model of signalling. This allows the assessment of inferred networks and predicted trajectories against a true gold standard.
A computational signalling model was used to generate time-courses of phosphoprotein abundance levels. The model describes the biochemistry underlying a realistic signalling network. Data were generated for combinations of 2 ligand stimuli (each one at $2$ concentrations, denoted to as ``lo'' and ``hi'') and $3$ inhibitors, or no inhibitor (as for the experimental data described above, cells were pre-incubated with the inhibitor prior to ligand stimulation). For each condition, a time-course of $20$ phosphoprotein levels is provided at $10$ time points post-stimulus.
It must be noticed that phosphoprotein names have been anonymized so that detailed prior information from canonical signalling pathways cannot be used. Efforts have been made to model the antibody-based readout of the RPPA platform and its technical variability in a faithful manner. Three technical replicates are provided per condition. Data provided to participants are protein abundance measurements on a linear scale. In this task, a single network should be inferred in contrast to the proteomic data challenge that requires 32 networks.



Following the approach used to evaluate the results submitted to the challenge, we have considered the same method to assess the performance of our predictions.
More precisely, in real data, for any given context, the set of nodes that showed salient changes under a test inhibitor (here an mTOR inhibitor) relative to the control was identified. These ``gold-standard'' sets are derived from (held-out) experimental data and should not be regarded as representing a fully definitive ground truth. For each predicted network, the set of mTOR descendants is predicted and compared against the experimental one to obtain the area under the receiver operating characteristic curve (AUROC) score \citep{hill_natmet_2016}. 
Results are ranked in each of the $32$ contexts by AUROC score, and the mean rank across contexts was used to provide an overall score and a final ranking. For the in silico data task, the true causal network was known and it was used to obtain an AUROC score for each predicted network. This score has been considered to determine the final ranking.

\begin{figure}[t]
\includegraphics[width=0.4\textwidth]{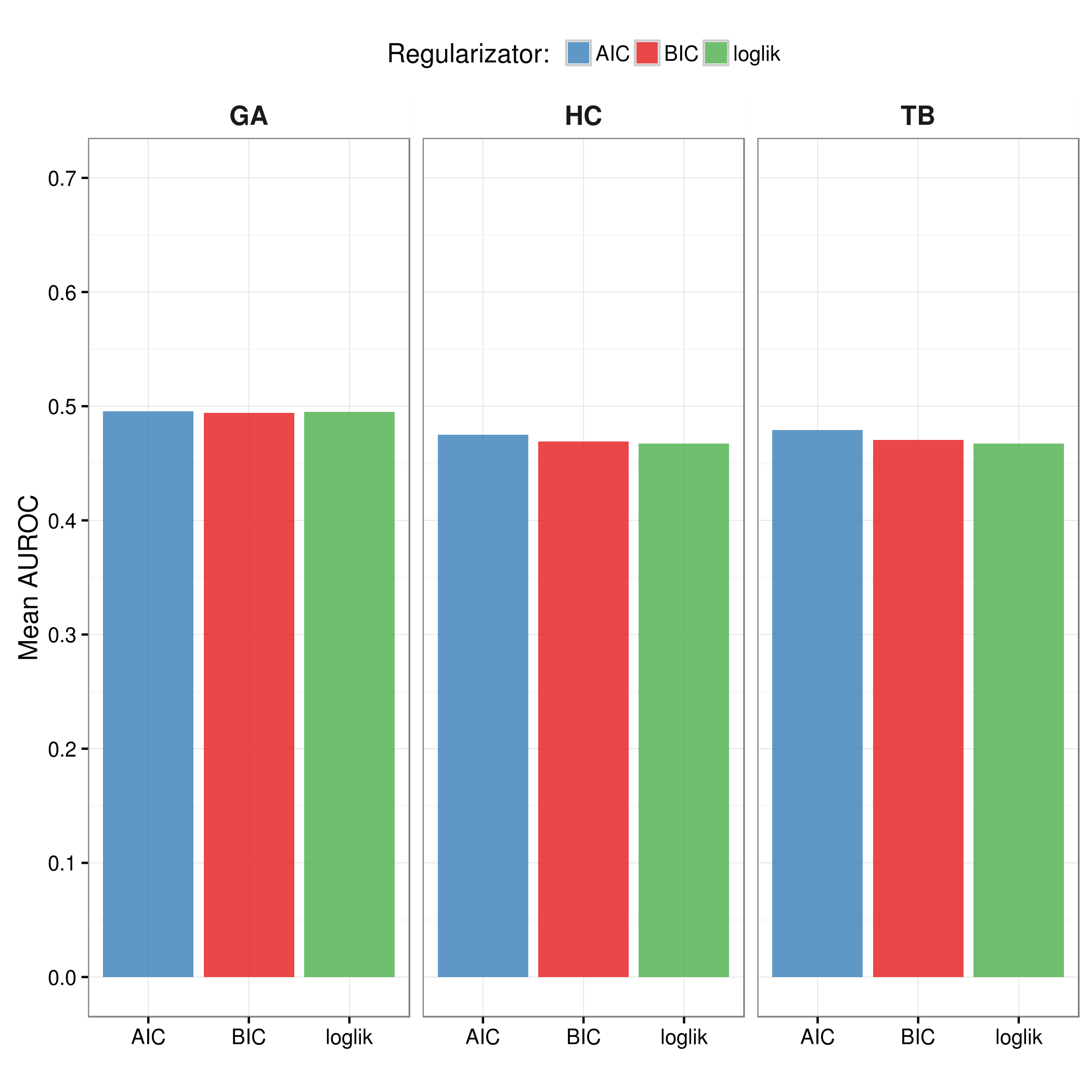}
\caption{Mean results on the $32$ experimental datasets for the considered approaches.}
\label{fig:hist}
\end{figure}

\begin{figure*}[!ht]
	\begin{center}
		\begin{tabular}{cc}
			\includegraphics[width=0.4\textwidth]{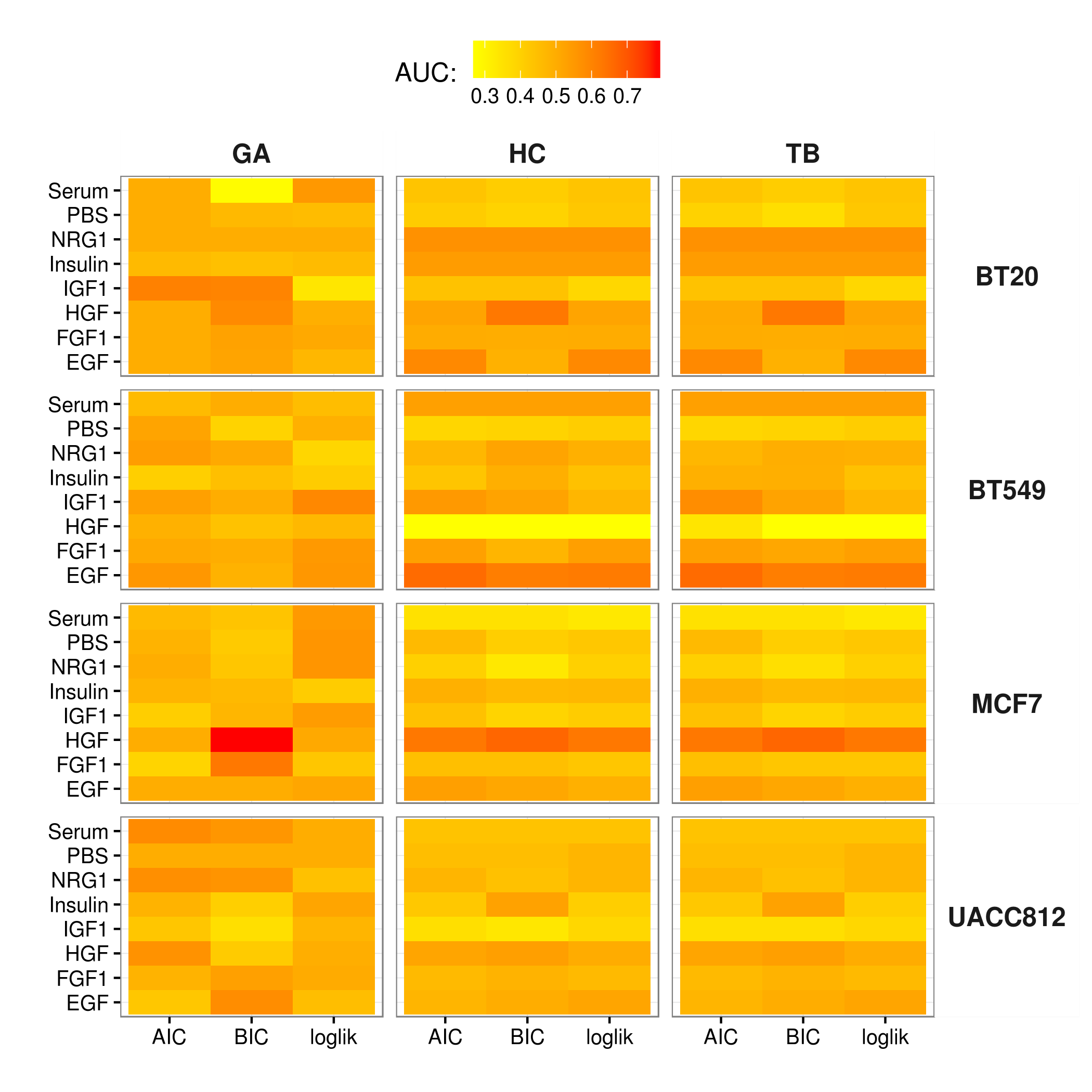} &
			\includegraphics[width=0.4\textwidth]{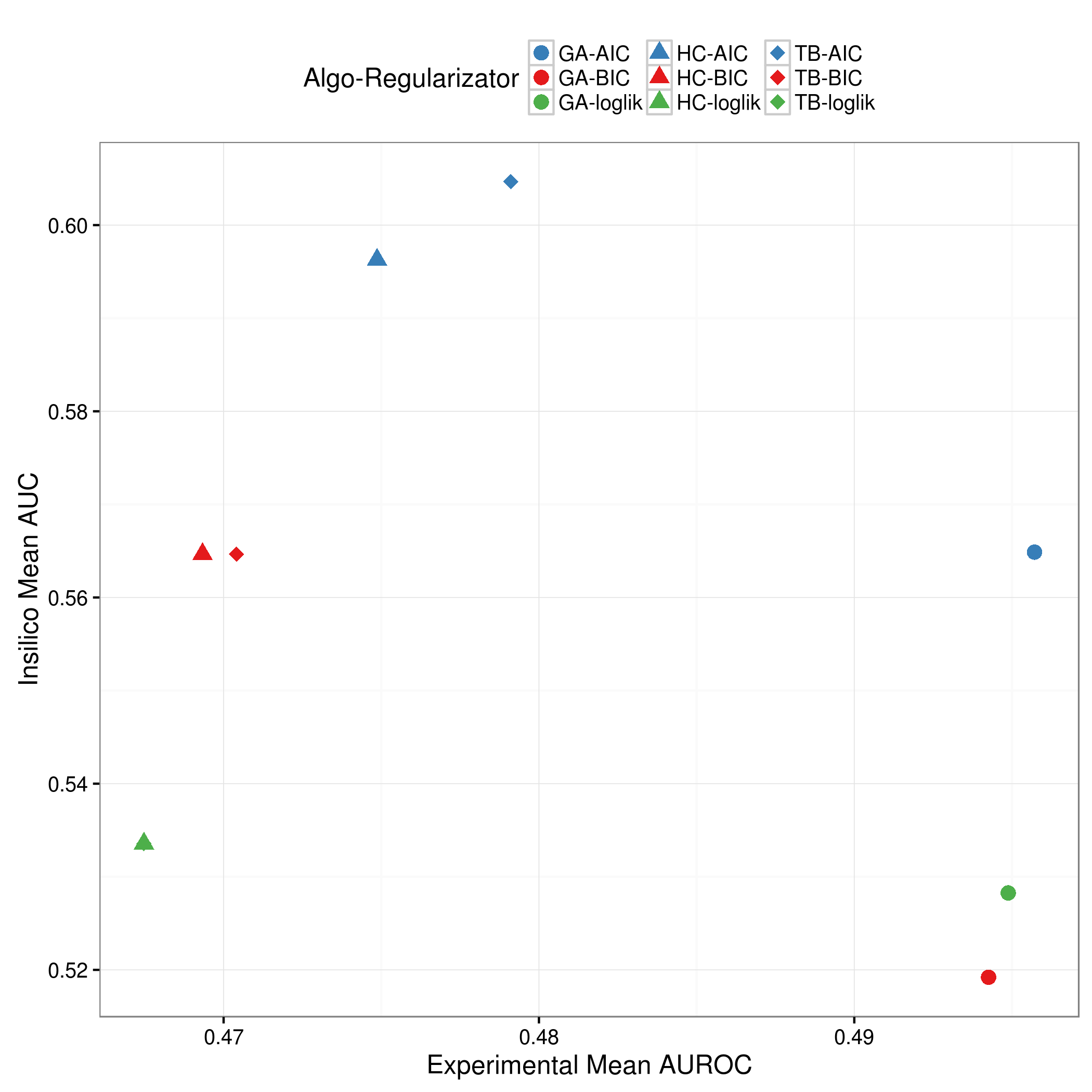} \\
			(a) & (b) \\
		\end{tabular}
	\end{center}
	\caption{(a)Heatmap showing the AUC values obtained with each combination of heuristic search method and regularizator on the $32$ experimental datasets. (b) Scores in experimental and in silico data tasks. Each combination of shape/color corresponds to a specific algorithm/regularizator pair.}
	\label{fig:scatter}
\end{figure*}

%

By analysing the mean AUROC values computed on the predictions on the $32$ real datasets, which are reported as bars in the plots in Figure~\ref{fig:hist}, it is possible to observe that all the tested approaches have similar performance, with mean values around $0.5$.

Anyway, when looking more in details on each of the $32$ datasets, we can draw more accurate considerations about the behaviour of the tested techniques. In particular, as showed in the heatmap in Figure~\ref{fig:scatter}(a), on the processed datasets we have obtained AUROC values ranging from $0.3$ to $0.7$.
As corroborated by several studies present in literature, these results highlight the fact that HC (hill climbing) and TB (\emph{tabu} search) have almost the same behaviour, also, w.r.t.~the considered regularizator, on the majority of the datasets.
On the other hand, GA (genetic algorithm) presents slightly different results than those obtained by the other two methods and, moreover, it seems that the results are affected by the considered regularizator.
Interestingly, when looking at the in silico AUC values, we can observe that, for each regularizator, HC and TB perform better on the in silico dataset, while GA is slightly worse; the opposite situation is observed in the real datasets, where the latter method (i.e.~genetic algorithms) achieves better results with respect to the two former techniques (i.e.,~hill climbing and \emph{tabu} search).
The scatter plot in Figure~\ref{fig:scatter}(b) shows a comparison of the mean AUC results on the in silico dataset against the AUROC mean values on the real datasets obtained with all the employed approaches.

To assess the quality of the obtained results, we performed a comparison with those obtained by the participants of the challenge.
More precisely, as reported in \citep{hill_natmet_2016}, several different techniques have been used to reconstruct the network proposed in this challenge, which can be distinguished based on the fact that a prior knowledge has been employed in order to improve the predictions, and also based on the reconstruction method (Bayesian networks in our case).
From the results on the in silico dataset, ranked by the mean AUC, we observed that our best performer (TB with AIC) obtained a value of $0.6$, which is better than all the other methods based on Bayesian networks and ranks in the top $15\%$ of the overall evaluated techniques.
On the other hand, on the $32$ real datasets our results are similar to those obtained by methods based on Bayesian networks, which present values around $0.5$.
Both these results are not surprising, since we do not use any prior knowledge on the input data (resulting in good performance on the in silico dataset), and also the number of observations in each of the $32$ real datasets is quite low compared to the number of nodes (phosphoprotein) of the networks to reconstruct, hence penalizing Bayesian approaches, making the inference task difficult.

\section{\uppercase{Conclusions}}
\label{sec:concl}
\noindent
In this work, we studied the inference of causal molecular networks, specifically focusing on signaling downstream of receptor tyrosine kinases. We modeled relationships (edges) in causal molecular networks ('causal edges') as directed links between nodes, in which inhibition of the parent node can lead to a change in the abundance of the child node, either by direct interaction or via unmeasured intermediate nodes.

To this extent, we have tested different methods to reconstruct (Bayesian) networks on real and in silico datasets proposed in the HPN-DREAM challenge.
Specifically, we analyzed the performance of different optimization search schemes, i.e., Hill climbing (HC), Tabu seach (TS) and Genetic algorithms (GA), and various likelihood scores, i.e., loglik, AIC and BIC. This analysis seems to show a better performance of more sophisticated search strategies like GA on real datasets, even if on in silico data it is shown that easier search schemes as HC and TS also prove to be very effective.

Furthermore, we find the obtained results to be encouraging, especially considering the fact the we have employed ``standard'' versions of the algorithms for the reconstruction of the network without making use of any biological prior.
%

%
%
\balance
\bibliographystyle{apalike}
\bibliography{biblio}

\end{document}